\documentclass{article}

\usepackage{arxiv}

\usepackage[utf8]{inputenc} 
\usepackage[T1]{fontenc}    
\usepackage{hyperref}       
\usepackage{url}            
\usepackage{booktabs}       
\usepackage{amsfonts}       
\usepackage{nicefrac}       
\usepackage{microtype}      
\usepackage{lipsum}
\usepackage{graphicx}
\graphicspath{ {./images/} }

\title{Image to Image Translation : Generating maps from Satellite Images}

\author{
 Vaishali Ingale \\
  Department of Information Technology\\
  Army Institute of Technology, Pune\\
  \texttt{vingale@aitpune.edu.in} \\
   \And
 Rishabh Singh \\
  Department of Information Technology\\
  Army Institute of Technology, Pune\\
  \texttt{rishabsingh\textunderscore17423@aitpune.edu.in} \\
  \And
 Pragati Patwal\\
  Department of Information Technology\\
  Army Institute of Technology, Pune\\
  \texttt{pragatipatwal\textunderscore17453@aitpune.edu.in} \\
}

\begin{document}
\maketitle
\begin{abstract}
Generation of maps from satellite images is conventionally done by a range of tools. Maps became an important part of life whose conversion from satellite images may be a bit expensive but Generative models can pander to this challenge. These models aims at finding the patterns between the input and output image. Image to image translation is employed to convert satellite image to corresponding map. Different techniques for image to image translations like Generative adversarial network, Conditional adversarial networks and Co-Variational Auto encoders are used to generate the corresponding human-readable maps for that region, which takes a satellite image at a given zoom level as its input.\\
We are training our model on Conditional Generative Adversarial Network which comprises of Generator model which which generates fake images while the discriminator tries to classify the image as real or fake and both these models are trained synchronously in adversarial manner where both try to fool each other and result in enhancing model performance. 
\end{abstract}



\section{Introduction}
Map creation is very costly and time consuming process; having a wide range of applications with utmost importance[1]. In several sectors of economy, maps have economic value for businesses: Driverless car companies (like Tesla), food delivery companies (like zomato, food Panda), ride sharing companies (like Ola and Uber), E-commerce companies (like ebay, Amazon, Alibaba), military and security agencies and a lot of other sectors.[1] Currently satellites, GPS sensors mounted drones or UAVs and a variety of reliable organizations are used for collecting geo-spatial data which is then organized together to form a detailed digital image (map). But the actual geographic conditions/street views and the publicly available human-readable maps have a significant amount of latency associated with them.\\ 
Automating this process of converting a satellite image into a human-readable map is one way to reduce this latency. This automation can be achieved by using Generative models. For training purpose pix2pix dataset has been used from kaggle which consist of 1096 concatenated satellite image with its corresponding map image. The authors of [2] has made the dataset publicly available and this dataset is obtained from Google Earth Engine dataset for satellite images and Google Maps API for corresponding map image. The satellite images are at a specified zoom level and resolution along with their corresponding human readable maps. The satellite image is passed to the generator of Conditional adversarial network as input which further generates an output image(fake). Both satellite image and output image are then paired and passed to the discriminator as input which further classifies images as Real or fake. Both the models are trained synchronously in adversarial manner where both try to fool each other and result in enhancing model performance.

\section{RELATED WORK}
A lot of significant work has already been done in the field of image-to-image translation using generative models. It can be categorized broadly into machine learning based algorithms and deep learning algorithms.\\
The former does not perform well on large datasets but the latter is very effective for image-to-image translation. It includes Variational Auto-encoder [6] and Generative adversarial networks [7]. The performance of both the models is comparable but GANs outperforms the variational Auto-encoder model.  After it was first described in 2014 [7], use of GAN has been increased significantly. Its derivative Conditional GAN [8],[3] is an improvised model for image-to-image translation conditioned on the target. [11] gives the basic formulation for training a classifier with the help of a GAN. Recent work also includes generation of high-resolution customizable portraits trained on massive corpora using styleGAN [10]. The recently booming Cycle-GAN [9] is also used for training unsupervised models using unpaired collection of images.\\
Using Cycle-GAN, we don't have any control over the output image generated by the generators, thus creating noise in the process.[2] proposed the modification in conditional GAN architecture to get an appropriate model for the generating map image from its corresponding satellite image as input. Authors of [2] used U-NET architecture for generator and Convolutional Patch-GAN classifier for discriminator. In this project we will use conditional GAN architecture with few modifications and GAN tricks to train our model.

\section{TECHNICAL APPROACH}
Fig. 1 is the flow diagram of the system containing all the steps involved in the experiment.\\
The dataset contain two folders: train folder containing 1,097 images and a validation folder containing 1,099 images.The images are loaded and re-scaled. And further splitted into the satellite image and Google map image. The source image is fed to the generator conditioned on target image to generate an output image. The discriminator is fed both with the output of generator and the expected image and predicts whether generated image is real or fake. The GAN objective loss function updates the discriminator and generator.
\begin{figure}[htbp]
\centerline{\includegraphics[width=9cm,height=12cm,keepaspectratio]{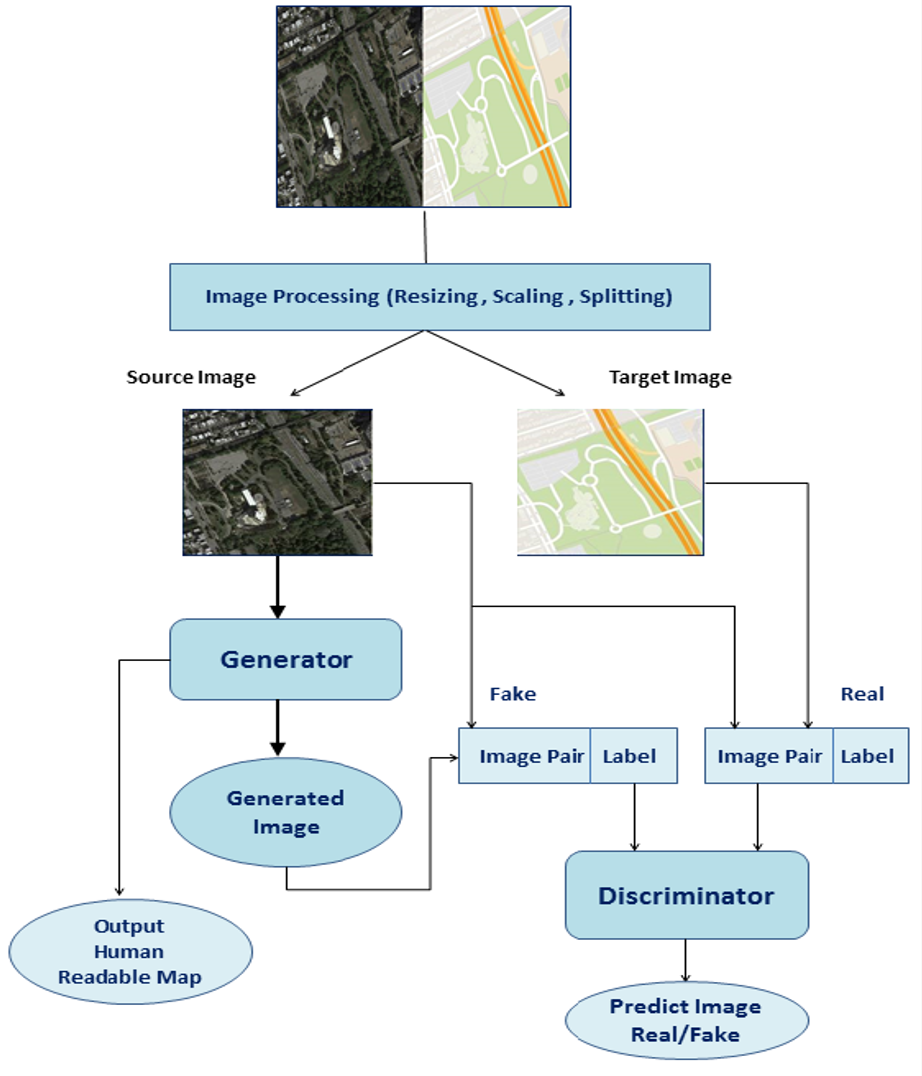}}
\caption{Technical Approach}
\label{fig1}
\end{figure}

\section{METHOD}
\label{sec:headings}
In Conditional GAN the generator is fed with latent space vector extracted from the satellite image, to generate the map image. Latent space vector is a compressed depiction of the satellite image used to interpolate the information and create new maps.\\
The Generator in CGAN is trained to generate a fake image which tries to fool the Discriminator and this adversarially trained discriminator try to distinguish between the real and fake sample.

\subsection{Objective Function}
GANs objective loss function is also called min-max loss as the generator and the discriminator simultaneously try to optimize the loss function by minimizing the generator's loss along with maximizing the discriminator's loss.\\
Loss Function of GAN can be further categorized as:\\
\subsubsection{Discriminator loss}
When the Discriminator network is trained, it try to classifies images generated by Generator network as Real or Fake and whenever it fail to do so, it penalizes itself for misclassifying Real as fake or Fake(generated by Generator) as Real by given function.\\
$$\hspace{2cm}max E_{\textbf{x} ~ \textbf{p}_{Real}} [log D (\textbf{x})] + E_{\textbf{z} ~ \textbf{p}_{Noise}} [log(1 \textendash D(G(\textbf{z})))]$$ \\

\begin{itemize}
    \item log(D({\textbf{x}})) is probability where Discriminator is correctly classifying generated image from real image.
    
    \item Maximizing log(1 \textendash D(G({\textbf{z}}))) help the Discriminator to correctly label sample generated by the Generator network as fake.
\end{itemize}

\vspace{0.6cm}

\subsubsection{Generator Loss}
When the Generator network is trained, input image is fed to the Generator and it Generates a image, then generated image to pass to discriminator which classified it as Real or Fake.\\
The generator uses discriminator to calculate its loss and it penalizes the Discriminator whenever it correctly classifies the generated image using given function.\\

$$ \hspace{2cm}min E_{ \textbf{z} ~ \textbf{p}_{Noise}}[log(1 \textendash D(G(\textbf{z}))] $$
\\

\subsection{Network Architecture}
The network architecture consists of two model : A Generator Network and A Discriminator Network where the Generator model generates a map image for corresponding satellite image and Discriminator model classifies whether generated image is real or fake.
\subsubsection{Generator Network}
Generator is a neural network model consisting of series of encoder and decoder blocks using U-Net architecture. Here Encoder and Decoder are using convolutional layers, dropouts, batch{\textendash}normalisation and activation layers.
This model take Satellite image as a input and generates Human readable map as output. Here what it does is, it down-sample the satellite image into its bottle neck representation then up-sample it from bottle neck (vector space representation) into the size of output image which is same as input size.
Generator has U-Net architecture which means it has skips connections that is every (i)th encoder block is connected to (n-i)th decoder block.\\
Fig. 2 refers to U-Net Generator Model Architecture.

\vspace{0.6cm}
\begin{figure}[htbp]
\centerline{\includegraphics[width=9cm,height=20cm,keepaspectratio]{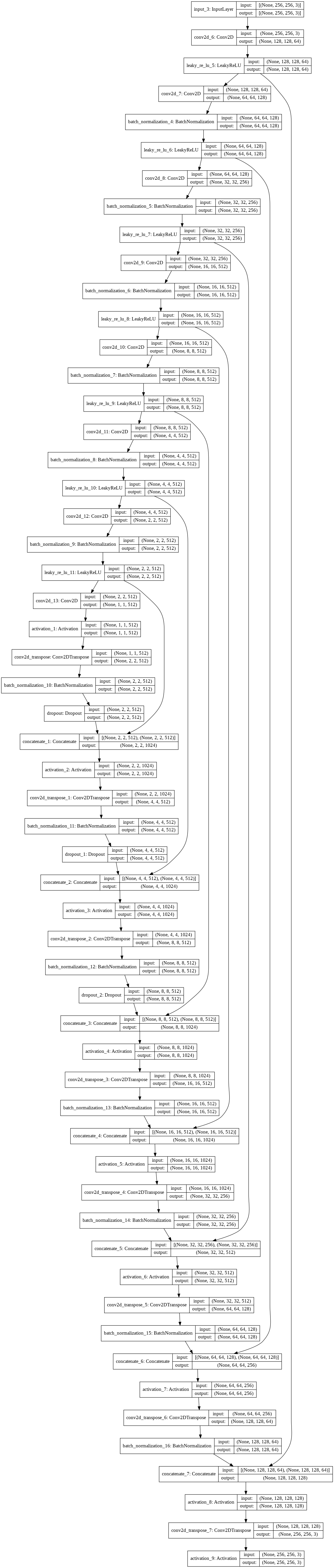}}
\caption{Generator Model}
\label{fig2}
\end{figure}

\subsubsection{Discriminator Network}
Discriminator model is deep convolutional neural network model which simply perform image classification. It is fed with the fake image pair (satellite image and generated map) and predicts whether this pair is real or fake. \\
Fig. 3 refers to the Architecture of Discriminator Model.

\begin{figure}[htbp]
\centerline{\includegraphics[width=9cm,height=20cm,keepaspectratio]{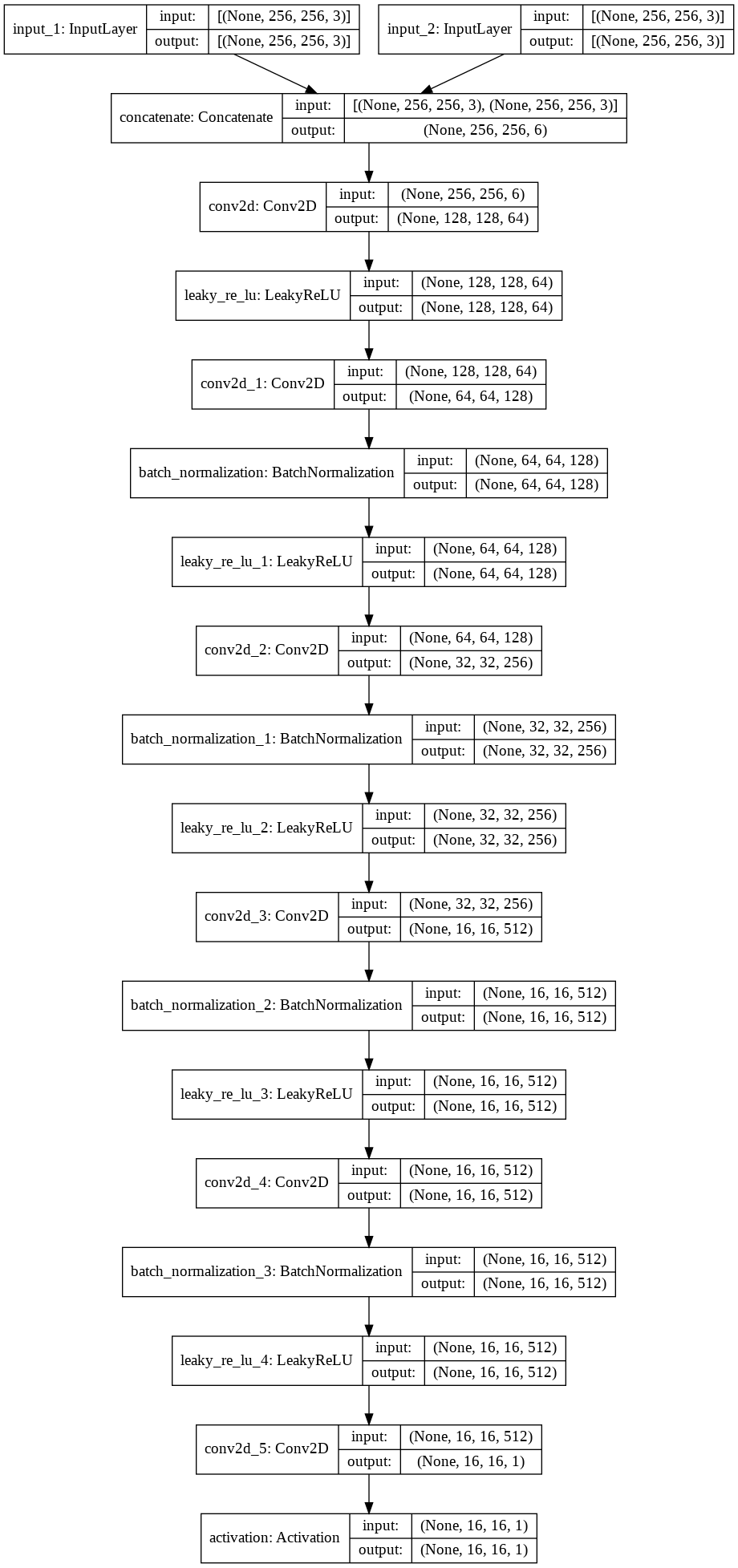}}
\caption{Discriminator Model}
\label{fig3}
\end{figure}

\newpage
\subsection{Result}
The training was done on dataset containing 1097 training images and batch size of 10 images is passed per epoch.\\
First Column contains the satellite images, second and third column contains the generated and the real map images for the given satellite image.

\begin{figure}[h]
			\centering
			\includegraphics[width=4.00in]{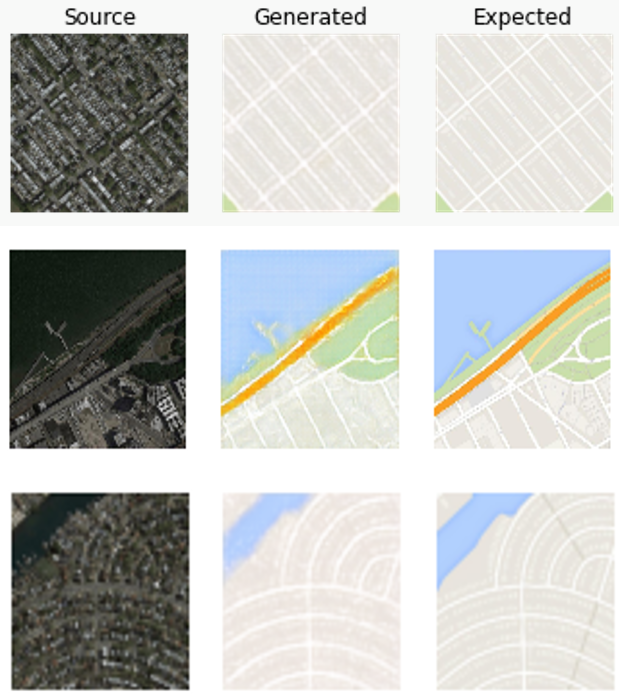}
			\label{fig_13}
	\end{figure}
    

\section{CONCLUSION}
We have implemented a feasible model which converts a satellite image to its corresponding map image. The result has been obtained by using few GAN tricks which includes normalizing the image between -1 to 1, using tanh as the last layer of generator output, batch normalization for GAN training and by avoiding sparse gradient such as RELU and maxpool to achieve GAN stability.The results can be improved by increasing the training time.\\ 
Future work also includes training on more bands, such as Infrared or panchromatic bands which could help in distinguishing more detailed features of the image such as lakes, vegetation slopes, shorelines etc.

\end{document}